 \documentclass[tablecaption=bottom,wcp]{jmlr} % W&CP article

\usepackage{booktabs}
\usepackage[load-configurations=version-1]{siunitx} % newer version
\usepackage{caption}
\theorembodyfont{\upshape}
\theoremheaderfont{\scshape}
\theorempostheader{:}
\theoremsep{\newline}

\jmlrproceedings{AABI 2018}{1st Symposium on Advances in Approximate Bayesian Inference, 2018}
\usepackage{todonotes}
\usepackage{dsfont}
\newcommand{\cF}{ {\cal F} }

\newcommand{\cL}{ {\cal L} }
\newcommand{\vU}{ \mathbf{U} }
\newcommand{\order}{\mathcal{O}}
\newcommand{\XX}{ {\cal X} }
\newcommand{\NN}{ {\cal N} }
\newcommand{\RR}{ \mathds{R} }
\newcommand{\msum}{\textstyle \sum}
\newcommand{\EE}{ \mathds{E} }
\newcommand{\given}{\,|\,}
\newcommand{\KL}{\operatorname{KL}}
\newcommand{\trace}{\operatorname{tr}}
\newcommand{\dee}{\mathrm{d}}
\newcommand{\transposed}{^\intercal}
\newcommand{\diag}{\operatorname{diag}}

\title[Sparse Variational GAM]{Scalable GAM using sparse variational Gaussian processes}

  \author{\Name{Vincent Adam} \Email{vincent.adam@prowler.io}\\\and
   \Name{Nicolas Durrande} \Email{nicolas@prowler.io}\\\and
   \Name{ST John} \Email{st@prowler.io}\\
   \addr PROWLER.io, 72 Hills Road, Cambridge, CB2 1LA, United Kingdom}

\begin{document}

\maketitle

\begin{abstract}
Generalized additive models (GAMs) are a widely used class of models of interest to statisticians as they provide a flexible way to design interpretable models of data beyond linear models. We here propose a scalable and well-calibrated Bayesian treatment of GAMs using Gaussian processes (GPs) and leveraging recent advances in variational inference. We use sparse GPs to represent each component and exploit the additive structure of the model to efficiently represent a Gaussian \emph{a posteriori} coupling between the components.
\end{abstract}

\section{Introduction}
\label{sec:intro}

Generalized additive models (GAMs) are a class of interpretable regression models with non-linear yet additive predictors \citep{hastie2017generalized}.
Their Bayesian treatment requires the specification of priors over functions. Here, we use Gaussian processes (GPs) \citep{rasmussen2006gaussian} and propose an approximate inference algorithm that is scalable with both the number of data points and additive components and that provides accurate posterior uncertainty estimates. 
We extend the variational pseudo-point GP approximation \citep{titsias2009variational, bauer2016understanding} to  posterior dependencies across GPs. This approximation provides state-of-the art performance for GP regression and provides approximations to the posterior distributions in the form of a GP. This approach has been successfully extended to the multiple GP setting using a factorized (mean-field) approximation of the posterior across GPs \citep{saul2016chained,adam2016scalable}. However, it suffers from the known variance underestimation of mean-field approximations and therefore can lead to poor predictions or can bias learning \citep{turner2011two}. \citet{adam2017structured} introduced additional structure to the posterior distribution by allowing some coupling across the inducing variables of the different GPs but this was at the cost of scalability.

\section{Background}
\subsection{Regression with multiple GPs}

We consider models with additive predictor and factorizing likelihood $p(Y\given f_{1\dots C}, X)=\prod_{n=1}^N p(y_{n} \given \msum_c f_c(x_{n}))$, where $f_1,\dots,f_C$ are functions from $\XX_c \to \RR$. The specific form of the likelihood is arbitrary. 
We denote $\cF=\{f_{1},...,f_{C}\}$ such that $p( \cF ) = \prod_c p(f_c)$ constitutes the joint distribution over the \emph{a priori} independent processes. 
$\cF (x)= [ f_{1}(x),...,f_{C}(x)]$
 is the vector of function evaluations at $x$. To simplify notation, when no argument is given, $\cF = \cF(X) \in \RR^{NC}$.  We denote by $K_{\cF,\cF}$ the block-diagonal prior covariance matrix over $\cF$. We are interested in computing the joint posterior $p( \cF \given X,Y)$. 

\subsection{Variational Inference}

The classical variational lower bound (or ELBO) to the marginal likelihood is given by  
\begin{equation} \label{eq:bound1}
\log p(Y\given X) \geq \EE_{q(\cF)}[ \log p(Y \given \cF, X) ] - \KL[q(\cF) \,\|\, p(\cF)] = \cL(q).
\end{equation}
This is the optimization objective in the Variational Free Energy (VFE) approximation.
We choose $q(\cF)$ to be a multivariate normal distribution with mean $\mu_{\cF}$ and covariance $\Sigma_{\cF}$, which is not an approximation in the conjugate likelihood setting. This leads to 
\begin{equation} \label{eq:bound1}
\cL(q) = \EE_{q(\cF)} [\log p(Y \given \cF)] + \frac{1}{2}\Big( \trace( K_{\cF,\cF}^{-1} \Sigma_{\cF} ) +  \mu_{\cF}\transposed K_{\cF,\cF}^{-1} \mu_{\cF}
 - \log |\Sigma_{\cF}| 
 - NC \Big).
\end{equation}
The expectation term in equation \eqref{eq:bound1} is intractable in most cases and needs to be approximated. See \cite{hensman2015scalable} for deterministic approximations and \cite{salimbeni2017doubly} for stochastic approximations.

\section{Optimal Gaussian posterior in variational inference}

Following \cite{opper2009variational}, we derive the expression for the optimal $\Sigma_{\cF}$ by noting that at the optimum, $\nabla_{\Sigma_{\cF}} \cL(q) = 0$. This implies that 
\begin{equation} \label{eq:opt_general}
\Sigma^{-1}_{\cF} =   K_{\cF,\cF}^{-1} + \nabla_{\Sigma_{\cF}}\big[\EE_{q(\cF)} [\log p(Y \given \cF)] \big].
\end{equation}

\subsection{Optimality in the additive case}

In the additive case considered here, the gradient term in  \eqref{eq:opt_general} is low-rank and can be parameterized by a vector $\lambda \in \RR^N$ as follows, with $\Lambda = \diag(\lambda)$ and $1_C = [\underbrace{1,\dots,1}_{C\text{ times}}]\transposed$:
\vspace*{-0.5cm}
\begin{equation} \label{eq:opt_add}
\Sigma^{-1}_{\cF} =   K_{\cF,\cF}^{-1} + (1_C \otimes \Lambda)(1_C \otimes \Lambda)\transposed.
\end{equation}
This parameterization requires $2N$ values, equal to that of the classical single GP regression setting described in \cite{opper2009variational}. It also inherits the non-convexity of this objective as highlighted by \cite{khan2012fast}.

\subsection{Optimality in the sparse additive case}

Following \cite{adam2016scalable} we introduce for each GP indexed by $c$ a set of $M$ `inducing points' $Z_{c}=[z_{c}^{(1)},...,z_{c}^{(M)}]\in \XX_{c}^{M}$. The vector of associated function evaluations is given by $\vU_{c}=[u_{c}^{(1)},...,u_{c}^{(M)}]=[f_{c}(z_{c}^{(1)}),...,f_{c}(z_{c}^{(M)})]$. We also define the stacked vector $\vU=[\vU_{1},...,\vU_{c}] \in \RR^{MC}$.

Following \cite{adam2017structured}, we parameterize $q(\cF) = q(\vU) \prod_{c} p(f_{c\neg\vU_{c}} \given \vU_{c})$.
This choice leads to a simplification of the lower bound \eqref{eq:bound1} as
\begin{align}
 \label{eq:bound2}
 \cL(q) &= \EE_{q(\cF)} [\log p(Y \given \cF,X)] - \KL[q(\vU) \,\|\, p(\vU)].
\end{align}

\cite{saul2016chained} considered the mean field case $q(\vU)=\prod_{c}q(\vU_{c})$ with each factor parameterized as a multivariate normal distribution $\NN(\mu_{\vU_c}, \Sigma_{\vU_c})$. This approach does not capture posterior dependencies across GPs.
\cite{adam2017structured} parameterized $q(\vU)$ as a multivariate normal distribution $\NN(\mu_{\vU}, \Sigma_{\vU})$ to include cross-GP coupling through the inducing variables $\vU$. We extend this last approach but ask what the optimal $q(\vU)$ should be. It turns out to be (see Appendix A):
\begin{align} \label{eq:optqU_sparse}
\Sigma^{-1}_{\vU,\vU} &=  K_{\vU,\vU}^{-1} + K_{\vU, \vU}^{-1}\big(\msum_c  K_{\vU, f_c} \big) \Lambda \big( \msum_{c'}K_{f_{c'},\vU} \big) K_{\vU, \vU}^{-1}.
\end{align}
This form has again $2N$ parameters which becomes an over-parameterization as soon as $N>M^2C^2/2$. Since we are interested in scalability, it is not of practical interest.

\section{A new parameterization for $q(\vU)$}

The second term of the sum in \eqref{eq:optqU_sparse} can be expressed as $AA\transposed$ with $A$ of size $MC \times N$. Keeping this structure arising from the additivity of the model, we propose the parameterization 
\begin{align}
 \label{eq:bound2}
\Sigma^{-1}_{\vU,\vU} &=   K_{\vU,\vU}^{-1} + BB\transposed,
\end{align}
with $B$ of size $MC \times M$ smaller than $A$. This parameterization preserves the \emph{structure} of the optimal covariance. It requires storing $M^2C$ values, which is less than a direct representation of a Cholesky factor of $\Sigma^{-1}_{\vU,\vU} $ that would require $M^2C^2$ parameters. 

\section{Summary of complexities}
Time and space complexity of the sparse variational algorithms are summarized in Table \ref{table:complexity}. 

\vspace*{-0.1cm}
\begin{table}[htbp]
 % The first argument is the label.
 % The caption goes in the second argument, and the table contents
 % go in the third argument.
\floatconts
  {table:complexity}%
  {\caption{Complexity of sparse variational additive models}}%
  {\begin{tabular}{llll}
  \bfseries Model & \bfseries Storage & \bfseries {\boldmath $\KL[q\,\|\,p]$} & \bfseries {\boldmath$\EE_{q(\rho)} \log  p(Y\given\rho)$} \\
  Mean field, \cite{saul2016chained} & $\order(CM^2)$ & $\order(CM^3)$ & $\order(CM^3 + NC M^2)$\\
  Coupled (covariance), \cite{adam2017structured} & $\order(C^2M^2)$ & $\order(C^3M^3)$ & $\order(C^3M^3 + N C^2M^2)$\\
  Coupled (precision), this work & $\order(CM^2)$ & $\order(CM^3)$ & $\order(CM^3 + NC^2M^2)$
  \end{tabular}\vspace{-2mm}}
\end{table}
\vspace*{-0.7cm}
\section{Related work}

Variational inference for the multi-GP setting has so far only used the mean-field (MF) approximation as described in \cite{saul2016chained}. 
When posterior dependencies are a quantity of interest, a natural approach is to increase the complexity of the variational posterior to capture these dependencies. This often results in a prohibitive increase in the complexity of the inference. 
Different solutions have been proposed to tackle this problem.
A first approach in \cite{giordano2015linear} consists of a two-step scheme where MF inference is \emph{assumed} to provide accurate posterior mean estimates. A perturbation analysis is then performed around the MF posterior means to provide second order (covariance) estimates.
A second approach consists in `relaxing' the MF approximation by extending the variational posterior $q(\cF)$ with additional multiplicative terms capturing dependencies while keeping the computational complexity of the resulting inference scheme low \citep{tran2015copula,hoffman2015stochastic}.
Our approach fits in this second family of extensions of the MF parameterization. It is tailored to the VFE approximation to GP models and leverages its sparsity to provide a fast and scalable inference algorithm.

\section{Illustration}

We consider a simple regression task consisting of approximating the following function: $f(x) = 10 \sin (\pi x_1 x_2) + 20(x_3 - 0.5)^2 +10x_4 + 5x_5$ with $x \in [0, 1]^6$ (note that the last variable has no effect), given 5000 observation points uniformly distributed in the input space and a Gaussian observation noise with unit variance.

\begin{figure}
    \centering
    \includegraphics[height=4.5cm]{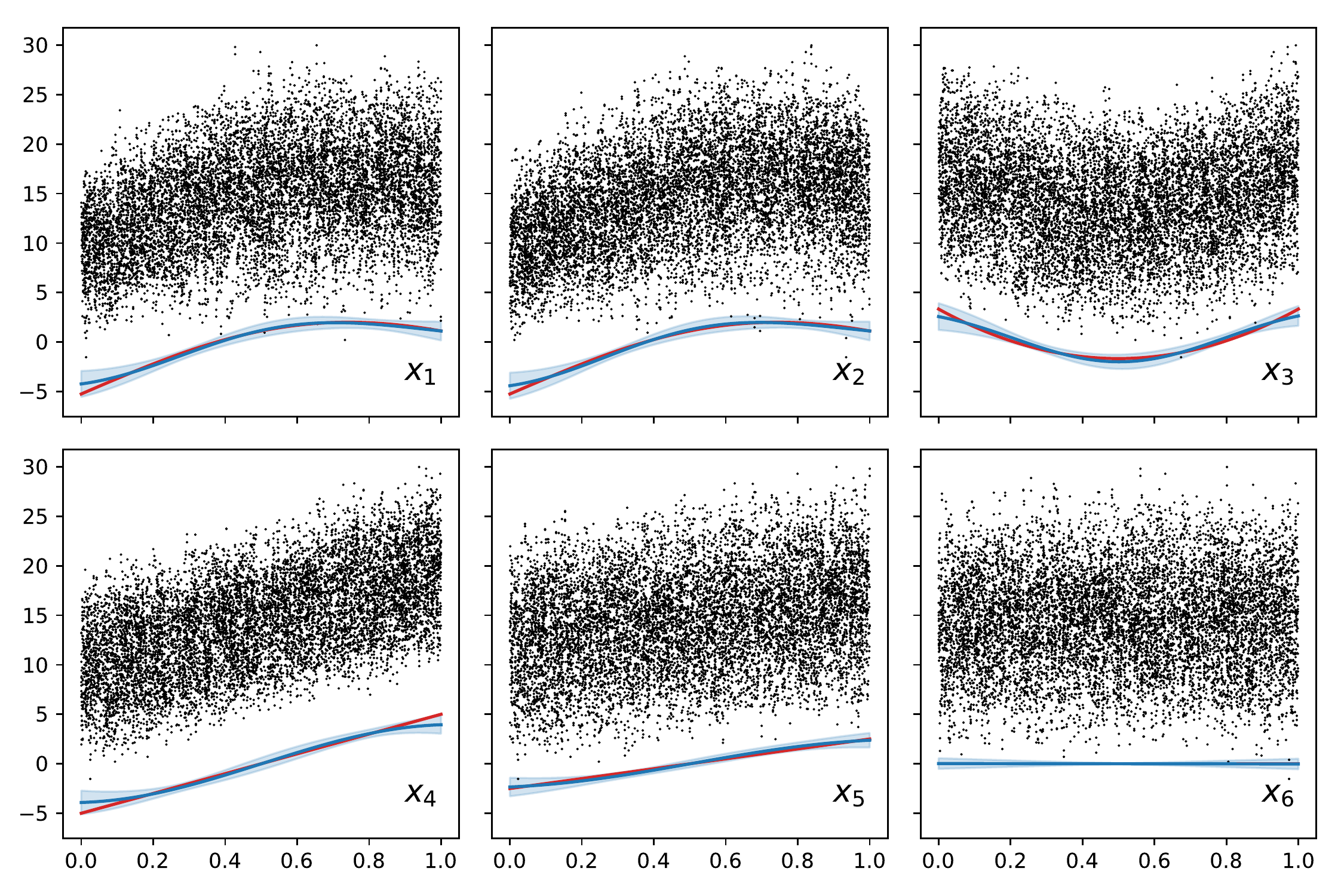}
    \includegraphics[height=4.5cm]{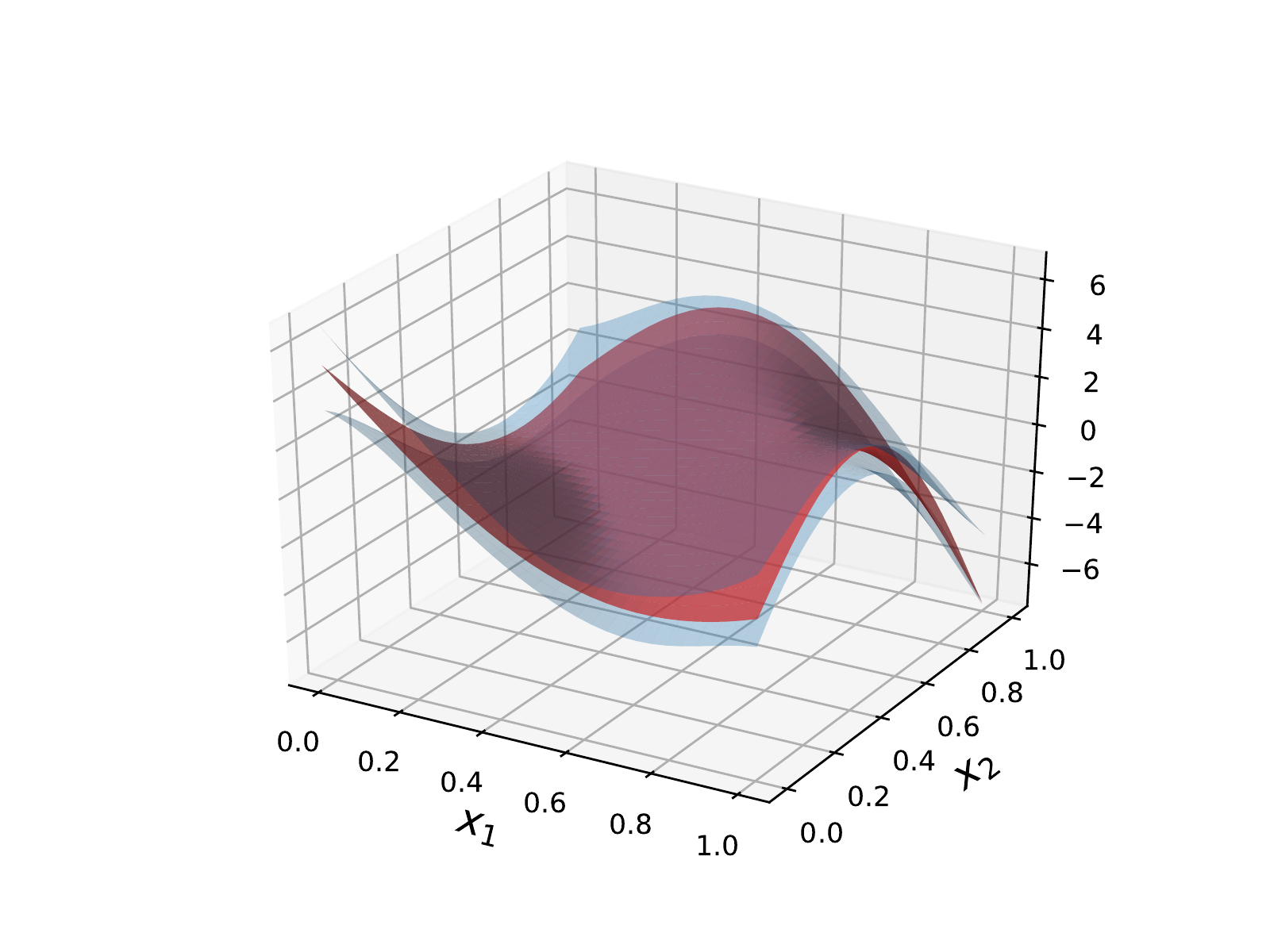}
    \caption{Left: projected observations (black points), estimated (blue) and analytically computed (red) univariate effects for each input dimension. Right: estimated (blue) and analytical (red) interaction between the variables $x_1$ and $x_2$.}
    \label{fig:anova}
\end{figure}

We choose a kernel dedicated to sensitivity analysis and tailored to the structure of the function at stake~[\cite{durrande2013anova}]. Given univariate squared exponential kernels $g_1, \dots, g_8$ we define the kernel as $k(x, y) = \sigma_0 + \sum_{i=1}^6 s_i(x_i, y_i) + s_7(x_1, y_1) s_8(x_2, y_2)$ with
\begin{align}
    s_i(x_i, y_i) = g_i(x_i, y_i) - \frac{ \int_0^1 g_i(x_i, s) \,\dee s \int_0^1 g_i(y_i, s) \,\dee s}{ \iint_0^1 g_i(s_i, t) \, \dee s \, \dee t} .
\end{align}
Since the number of observations is relatively large and the kernel has an additive structure (it is the sum of 8 kernels), we choose the \emph{sparse additive model} described above. We choose 16 regularly spaced one-dimensional inducing points for each kernel $s_1, \dots, \s_6$ and 16 points distributed as a $4 \times 4$ grid for the bi-dimensional kernel $s_7 s_8$. The final model is obtained by maximizing the ELBO with respect to the variational parameters and the hyper-parameters of the $g_i$. Given the structure of the model and the fact that inducing inputs are dedicated to model components, it is then possible to decompose the model predictions and to represent separately all the components of the ANOVA representation of the test function.
Figure~\ref{fig:anova} shows that the model accurately approximates the test function and that the proposed framework is helpful to reveal its inner structure. 

\section{Conclusion}

 We presented a method that provides a fast, scalable and well-calibrated Bayesian treatment of GAMs. Although motivated by GAMs, our structured variational distribution may be used in models where the predictor is non-additive but where the posterior is well-approximated by a unimodal distribution.
%\acks{Acknowledgements go here.}

\bibliography{main}

\appendix

\section{Optimal covariance in the additive case}\label{apd:first}

We first define $V(Y,\mu_{\cF}, \Sigma_{\cF}) = \EE_{q(\cF)} \log p(Y \given \cF) $. From \cite{opper2009variational}, we know that the optimal variational precision is structured as
\begin{equation*}
\Sigma^{-1}_{\cF} =   K_{\cF,\cF}^{-1} + \nabla_{\Sigma_{\cF}} V(Y,\mu_{\cF}, \Sigma_{\cF}).
\end{equation*}
For factorizing likelihood and additive predictors, and defining $\rho(\cdot) = \msum_c f_c(\cdot)$, we have
\begin{align*}
V(Y,\mu_{\cF}, \Sigma_{\cF}) &= \msum_n  \EE_{q(\rho(x_n) )} [\log p(y_n \given \rho(x_n)) ]\\
&= \msum_n  v(y_n, \mu_{\rho(x_n)},\sigma^2_{\rho(x_n)}) ,
 \end{align*}
where $q(\rho(x_n) )$ has variance $\sigma^2_{\rho(x_n)} = 1_C\transposed  \Sigma_{\cF(x_n)} 1_C = \msum_{c,c'}\Sigma_{f_c(x_n),f_{c'}(x_n)}$.\\
The gradient term in the optimal precision thus can be written as 
\begin{align*}
\nabla_{\Sigma_{\cF}} V(Y,\mu_{\cF}, \Sigma_{\cF}) &=  \msum_n   \frac{\partial }{\partial \sigma^2_{\rho(x_n)}}v(y_n, \mu_{\rho(x_n)},\sigma^2_{\rho(x_n)}) \nabla_{\Sigma_{\cF}}\sigma^2_{\rho(x_n)}\\
 &= \msum_n  \lambda_n^2 \msum_{cc'} e_{cn,c'n} ,
\end{align*}
where $e_{i,j}$ is the indicator matrix of size $NC \times NC$ with $1$ at location $(i,j)$.
With $\Lambda=\diag(\lambda)$, this can be rewritten in matrix form:
\begin{align*}
\nabla_{\Sigma_{\cF}}V(Y,\mu_{\cF}, \Sigma_{\cF}) &= (1 \otimes \Lambda)(1 \otimes \Lambda)\transposed .
\end{align*}

\section{ELBO evaluation: additive case}\label{apd:first}

We parameterize the approximate posterior as
\[
q(\cF) = \NN(K_{\cF,\cF}\,\alpha, \Sigma_{\cF}) 
\]
with
\[
\Sigma_{\cF} =\big(K_{\cF,\cF}^{-1} + (1 \otimes \Lambda)(1 \otimes \Lambda)\transposed\big)^{-1},
\]
and optimize 
\[
\cL(q) = \EE_{q(\sum_c f_c)} [\log p(y\given \msum_c f_c)] - \KL[q(\cF) \,\|\, p(\cF)],
\]
where
\[
\KL[q(\cF)\,\|\,p(\cF)] = \frac{1}{2}[- \log |K_{\cF,\cF}^{-1} \Sigma_{\cF}| + \alpha\transposed K_{\cF,\cF} \alpha + \trace(K_{\cF,\cF}^{-1}\Sigma_{\cF}) - NC ]. 
\]

\subsection{Computing marginals of $\Sigma_{\cF^{(n)}}$}

\begin{align*}
\Sigma_{\cF} &=\big(K_{\cF,\cF}^{-1} + (1 \otimes \Lambda)(1 \otimes \Lambda)\transposed\big)^{-1}\\
 &=K_{\cF,\cF} - K_{\cF,\cF} (1 \otimes \Lambda) \big(I + (1 \otimes \Lambda)\transposed K_{\cF,\cF} (1 \otimes \Lambda)\big)^{-1} (1 \otimes \Lambda)\transposed K_{\cF,\cF} \\
  &=K_{\cF,\cF} - K_{\cF,\cF} (1 \otimes \Lambda) A^{-1} (1 \otimes \Lambda)\transposed K_{\cF,\cF} ,
 \end{align*}
 where
 \begin{align*}
A &= I + (1 \otimes \Lambda)\transposed K_{\cF,\cF} (1 \otimes \Lambda)\\
 &= I + \msum_c  \Lambda\transposed K_{f_c,f_c} \Lambda .
 \end{align*}
To evaluate the ELBO we need, for each data point $(x_n,y_n)$, the marginal $q(\sum_c f^{n}_c)$.
This corresponds to the diagonal elements of
\begin{align*}
\Sigma_\text{sum} &= \!(1 \otimes I)\transposed \Sigma_{\cF} (1 \otimes I)\transposed\\
 &=\!(1 \otimes I)\transposed [ K_{\cF,\cF} - K_{\cF,\cF} (1 \otimes \Lambda) A^{-1} (1 \otimes \Lambda)\transposed K_{\cF,\cF} ] (1 \otimes I)\transposed\\
 &=\!\msum_c K_{f_c,f_c} -  \msum_{c,c'}  K_{f_c,f_c}  \Lambda A^{-1} \Lambda K_{f_{c'},f_{c'}}  \\
 &=\!\msum_c K_{f_c,f_c} -  \msum_{c,c'}  (L_A^{-\intercal} \Lambda K_{f_c,f_c})\transposed(L_A^{-\intercal} \Lambda K_{f_{c'},f_{c'}})  .
\end{align*}
\subsection{Computing the KL}
\begin{align*}
| K_{\cF,\cF}^{-1} \Sigma_{\cF} | &= |K_{\cF,\cF}^{-1}|/| \Sigma_{\cF}^{-1} |\\
 &=|K_{\cF,\cF}^{-1}|/|K_{\cF,\cF}^{-1} + (1 \otimes \Lambda)(1 \otimes \Lambda)\transposed|\\
 &=|K_{\cF,\cF}^{-1}|/[|I + (1 \otimes \Lambda)\transposed K_{\cF,\cF} (1 \otimes \Lambda)||I||K_{\cF,\cF}^{-1}|] \\
 &=1/|A|
\end{align*}
\begin{align*}
\trace( K_{\cF,\cF}^{-1} \Sigma_{\cF} ) 
&= \trace( K_{\cF,\cF}^{-1}(K_{\cF,\cF} - K_{\cF,\cF} (1 \otimes \Lambda) A^{-1} (1 \otimes \Lambda)\transposed K_{\cF,\cF}))\\
&= \trace( I -  (1 \otimes \Lambda) A^{-1} (1 \otimes \Lambda)\transposed K_{\cF,\cF})\\
&= NC -  \trace( A^{-1} (1 \otimes \Lambda)\transposed K_{\cF,\cF})\\
&= NC - \msum_c \trace(\Lambda A^{-1} \Lambda\transposed K_{f_c,f_c})\\
&= NC -  \trace(\Lambda A^{-1} \Lambda\transposed \msum_c K_{f_c,f_c})
 \end{align*}
In the end,
\[
\KL[q(\cF)\,\|\, p(\cF)] = \frac{1}{2}[ \log |A| + \alpha\transposed K_{\cF,\cF} \alpha -  \trace(\Lambda A^{-1} \Lambda\transposed \msum_c K_{f_c,f_c}) ].
\]

\subsection{Summary}

\fbox{
\centering
\begin{minipage}{\textwidth}

\begin{align*}
A &= I + \msum_c  \Lambda\transposed K_{f_c,f_c} \Lambda\\
\KL[q(\cF)\,\|\,p(\cF)] &= \frac{1}{2} [\log |A| + \alpha\transposed K_{\cF,\cF} \alpha - \trace(\Lambda A^{-1} \Lambda\transposed \msum_c K_{f_c,f_c}) ]\\
\mu_\text{sum} &=  \msum_c K_{f_c,f_c} \alpha_c \\
\Sigma_\text{sum} &= 
\msum_c \diag(K_{f_c,f_c}) -  \msum_{c,c'} \diag( K_{f_c,f_c}  \Lambda A^{-1} \Lambda K_{f_{c'},f_{c'}}) 
\end{align*}

\end{minipage}
}
\section{ELBO evaluation: sparse additive case}\label{apd:second}

We parameterize an approximate posterior over the inducing values as 
\[
q(\vU) = \NN(K_{\vU,\vU}\alpha, \Sigma_{\vU,\vU})
\]
with
\[
\Sigma_{\vU,\vU}=(K_{\vU,\vU}^{-1} + BB\transposed)^{-1},
\]
where $B = [B_1,\dots, B_C] \in \RR^{MC \times M}$.
We optimize 
\[
\cL(q) = \EE_{q(\sum_c f_c)} [\log p(Y \given \msum_c f_c) - \KL[q(\vU)\,\|\,p(\vU)]
\]
with 
\[
\KL[q(\vU)\,\|\,p(\vU)] = \frac{1}{2}[- \log |K_{\vU,\vU}^{-1} \Sigma_{\vU,\vU}| + \alpha\transposed K_{\vU,\vU} \alpha + \trace(K_{\vU,\vU}^{-1}\Sigma_{\vU,\vU}) - MC ].
\]

\subsection{Computing marginals of $\Sigma_{\cF}$}

We have
\begin{align*}
\Sigma_{\vU,\vU} &=(K_{\vU,\vU}^{-1} + B B\transposed)^{-1}\\
 &=K_{\vU,\vU} - K_{\vU,\vU} B (I + B\transposed K_{\vU,\vU} B)^{-1} B\transposed K_{\vU,\vU} \\
  &=K_{\vU,\vU} - K_{\vU,\vU} B A^{-1} B\transposed K_{\vU,\vU} 
\end{align*}
where $A = I + B\transposed K_{\cF,\cF} B$, so
\begin{align*}
\mu_{F} &= K_{\cF,\vU}K_{\vU,\vU}^{-1}K_{\vU,\vU}\alpha\\
 &= K_{\cF, \vU}\alpha
\end{align*}
and
\begin{align*}
\Sigma_{\cF} &= K_{\cF,\cF} - K_{\cF,\vU}(K_{\vU,\vU}^{-1} - K_{\vU,\vU}^{-1}\Sigma_{\vU,\vU} K_{\vU,\vU}^{-1})K_{\vU,\cF}\\
&= K_{\cF,\cF} - K_{\cF,\vU}(K_{\vU,\vU}^{-1} - K_{\vU,\vU}^{-1}[K_{\vU,\vU} - K_{\vU,\vU} B A^{-1} B\transposed K_{\vU,\vU}] K_{\vU,\vU}^{-1})K_{\vU,\cF}\\
&= K_{\cF,\cF} - K_{\cF,\vU}( B A^{-1} B\transposed )K_{\vU,\cF} .
\end{align*}
Therefore 
\begin{align*}
\Sigma_\text{sum} &=  \msum_c K_{f_c,f_c} - \msum_{c,c'} K_{f_{c},\vU}( B A^{-1} B\transposed )K_{\vU, f_{c'}} .\\
 &=  \msum_c K_{f_c,f_c} - \msum_{c,c'} K_{f_{c},\vU_c}( B_c A^{-1} B_{c'}\transposed )K_{\vU_{c'}, f_{c'}} .\\
 &=  \msum_c K_{f_c,f_c} - \msum_{c,c'} (L_A^{-1} B_c\transposed K_{\vU_c, f_{c'}})\transposed(L_A^{-1} B_{c'}\transposed K_{\vU_{c'}, f_{c'}}) .
\end{align*}
The Cholesky decomposition of $A=L_A L_A\transposed$  is of cost $\order(M^3)$. Solving operations  $L^{-1} B_c\transposed$ for each additive term costs a total of $\order(C M^{3})$. Computing the marginal predictor variances then costs an extra $\order(N C^2 M^2)$. In total, the computational cost of posterior predictions is $\order(C M^3 + N C^2 M^2)$.
\subsection{Computing the KL}
As in the additive case, we have
\[
| K_{\vU,\vU}^{-1} \Sigma_{\vU,\vU} | =1/|A|
\]
and
\begin{align*}
\trace( K_{\vU,\vU}^{-1} \Sigma_{\vU,\vU} ) 
&= \trace( K_{\vU,\vU}^{-1}(K_{\vU,\vU} - K_{\vU,\vU}B A^{-1} B\transposed K_{\vU,\vU}))\\
&= \trace( I - B A^{-1}B\transposed K_{\vU,\vU})\\
&= MC - \msum_c \trace(B_c A^{-1} B_c\transposed K_{\vU_c,\vU_c}) .
 \end{align*}
In the end,
\[
\KL[q(\vU)\,\|\,p(\vU)] = \frac{1}{2} [\log |A| + \alpha\transposed K_{\vU,\vU} \alpha - \msum_c \trace(B_c A^{-1} B_c\transposed K_{\vU_c,\vU_c}) ].
\]

\subsection{Summary}

\fbox{
\centering
\begin{minipage}{\textwidth}
\begin{align*}
A &= I + B\transposed K_{\cF,\cF} B\\
\KL[q(\vU)\,\|\,p(\vU)] &= \frac{1}{2}[ \log |A| + \alpha\transposed K_{\vU,\vU} \alpha - \msum_c \trace(B_c A^{-1} B_c\transposed K_{\vU_c,\vU_c}) ] \\
\mu_\text{sum} &=  \msum_c K_{f_c,\vU_c} \alpha_c \\
\Sigma_\text{sum} &= 
\msum_c \diag(K_{f_c,f_c}) -  \msum_{c,c'} \diag( K_{f_c,\vU_c}  B_c A^{-1} B_{c'}\transposed K_{\vU_{c'},f_{c'}})
\end{align*}
\end{minipage}
}
\end{document}